\documentclass[lettersize,journal]{IEEEtran}
\usepackage{amsmath,amsfonts}
\usepackage{algorithmic}
\usepackage{algorithm}
\usepackage{array}
\usepackage[caption=false,font=footnotesize,labelfont=rm,textfont=rm]{subfig}
\usepackage{textcomp}
\usepackage{stfloats}
\usepackage{url}
\usepackage{verbatim}
\usepackage{graphicx}
\usepackage{cite}
\usepackage{color}
\usepackage{bbding}
\usepackage{booktabs}
\usepackage{multirow}
\usepackage{xcolor}
\usepackage[export]{adjustbox}
\usepackage{hyperref}
\usepackage{cleveref}

\def\etal{\emph{et al.~}}
\newcommand{\hzw}[1]{\textcolor{black}{#1}}

\begin{document}

\title{Accurate and lightweight dehazing via multi-receptive-field non-local network and novel contrastive regularization}

\author{Zewei He, Zixuan Chen, Jinlei Li, Ziqian Lu, Xuecheng Sun, Hao Luo, \\Zhe-Ming Lu$^\ast$,~\IEEEmembership{Senior Member,~IEEE}, Evangelos K. Markakis,~\IEEEmembership{Member,~IEEE}
	\thanks{This work was supported in part by the National Natural Science Foundation of China under Grant No. 52305590, in part by the Zhejiang Provincial Natural Science Foundation of China under Grant No. LQ24F010004.}
	\thanks{Z. He is with Huanjiang Laboratory, Zhuji, P.R. China and School of Aeronautics and Astronautics, Zhejiang University, Hangzhou, P.R.China (e-mail: zeweihe@zju.edu.cn).}
	\thanks{Z. Chen, J. Li, Z. Lu, X. Sun, H. Luo and Z.-M. Lu are with School of Aeronautics and Astronautics, Zhejiang University, Hangzhou, P.R.China (e-mails: 22224039@zju.edu.cn, jinlei\_li@zju.edu.cn, ziqianlu@zju.edu.cn, xuechengsun@zju.edu.cn, luohao@zju.edu.cn).}
	\thanks{E. K. Markakis is with Electrical and Computer Engineering Department, Hellenic Mediterranean University, Heraklion, Crete, Greece (e-mail: Emarkakis@hmu.gr).}
	\thanks{$^\ast$Corresponding author: Zhe-Ming Lu (e-mail: zheminglu@zju.edu.cn).}
}

\markboth{Journal of \LaTeX\ Class Files,~Vol.~14, No.~8, August~2021}%
{Shell \MakeLowercase{\textit{et al.}}: A Sample Article Using IEEEtran.cls for IEEE Journals}

\IEEEpubid{0000--0000/00\$00.00~\copyright~2021 IEEE}

\maketitle

\begin{abstract}
Recently, deep learning-based methods have dominated image dehazing domain. 
A multi-receptive-field non-local network (MRFNLN) consisting of the multi-stream feature attention block (MSFAB) and the cross non-local block (CNLB) is presented in this paper to further enhance the performance. 
We start with extracting richer features for dehazing. 
Specifically, a multi-stream feature extraction (MSFE) sub-block, which contains three parallel convolutions with different receptive fields (i.e., $1\times 1$, $3\times 3$, $5\times 5$), is designed for extracting multi-scale features. 
Following MSFE, an attention sub-block is employed to make the model adaptively focus on important channels/regions. 
These two sub-blocks constitute our MSFAB. 
Then, we design a cross non-local block (CNLB), which can capture long-range dependencies beyond the query. 
Instead of the same input source of query branch, the key and value branches are enhanced by fusing more preceding features. 
CNLB is computation-friendly by leveraging a spatial pyramid down-sampling (SPDS) strategy to reduce the computation and memory consumption without sacrificing the performance. 
Last but not least, a novel detail-focused contrastive regularization (DFCR) is presented by emphasizing the low-level details and ignoring the high-level semantic information in a representation space \hzw{specially designed for dehazing}. 
Comprehensive experimental results demonstrate that the proposed MRFNLN model outperforms recent state-of-the-art dehazing methods with less than 1.5 Million parameters.

\end{abstract}



\section{Introduction}

\IEEEPARstart{I}{mages}
acquired under hazy conditions often exhibit significant degradation in visual quality, including reduced contrast and color distortion \cite{Tan2008CVPR}, which substantially undermines the performance of high-level vision tasks such as object detection and semantic segmentation \cite{Lin2023TMM, Wang2024TMM, Cheng2024TMM, Su2025TMM}. 
The demand for haze-free images is thus critical in these applications. 
Consequently, single image dehazing, a process aimed at restoring a clear scene from its hazy counterpart, has attracted considerable attention from both academic and industrial research communities over the past decade \cite{Cui2025TITS,Wang2023TITS}.

\begin{figure}
	\centering
	\includegraphics[width=0.98\linewidth]{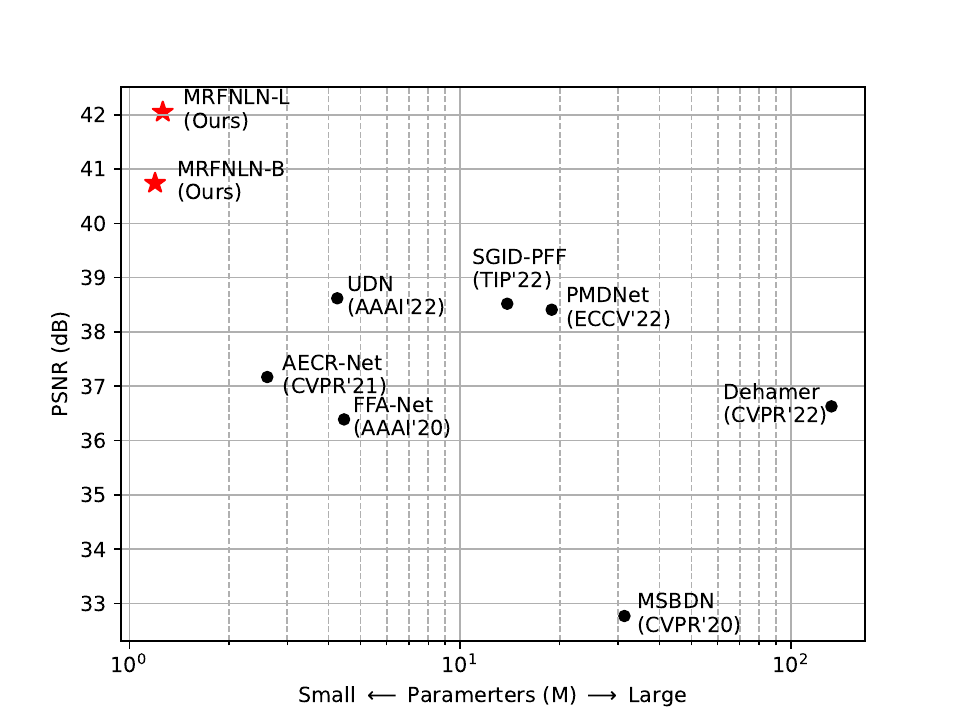}
	\caption{PSNR \emph{vs.} number of parameters. Compared with the state-of-the-art dehazing methods, our MRFNLN-B/-L can achieve highest PSNR value on SOTS-indoor dataset with significantly fewer parameters, indicating the efficiency and effectiveness.}
	\label{fig:fig0}
\end{figure}

\hzw{As a fundamental low-level image restoration task, the haze can be described by the Atmospheric Scattering Model (ASM) \cite{Nayar1999ICCV-ASM,Narasimhan2003TPAMI-ASM}:}
\begin{equation}
	I(x) = J(x)t(x)+A(1-t(x)),
	\label{Eqn:ASM}
\end{equation}
where $I$ denotes the observed hazy image, $J$ denotes the haze-free image, $A$ indicates the global atmospheric light describing the intensity of ambient light, $t$ represents the transmission map, and $x$ is the pixel coordinate.
\IEEEpubidadjcol

Given a hazy image, recovering its clean version is highly ill-posed.
Early approaches tend to solve this challenge by introducing various priors, such as Dark Channel Prior (DCP) \cite{He2009CVPR-DCP,He2011TPAMI-DCP}, Non-Local Prior (NLP) \cite{Berman2016}, Color Attenuation Prior (CAP) \cite{Zhu2015TIP}, etc.
These priors try to restrict the solution space to some extent.
However, haze removal quality relies heavily on the consistency between the introduced prior and the real data distribution.
The recovered image would be distorted/varicolored when the assumptions of these priors are not met.

In the past decade, convolutional neural networks (CNNs) has made a breakthrough, and many researchers have proposed numerous data-driven methods \cite{Cai2016TIP,Ren2016ECCV,Li2017ICCV-AOD,Zhang2018CVPR-DCPDN,Qin2020AAAI-FFA,Dong2020CVPR-MSBDN,Wu2021CVPR-AECR,Hong2022AAAI,Ye2022ECCV-PMDNet}.
Some of them employ CNN to estimate the $A$ and $t(x)$ in Eqn.~\ref{Eqn:ASM}, and then accordingly derive the haze-free prediction \cite{Cai2016TIP,Ren2016ECCV,Li2017ICCV-AOD,Zhang2018CVPR-DCPDN}.
The others directly learn the relationship between the hazy image and corresponding ground-truth to reconstruct the latent haze-free images (or haze residues) \cite{Qin2020AAAI-FFA,Dong2020CVPR-MSBDN,Wu2021CVPR-AECR,Hong2022AAAI,Ye2022ECCV-PMDNet}.
Normally, they try to improve the dehazing performance by increasing the depth and width of the networks.
However, the number of parameters and the training difficulty of such a model will substantially increase, as shown in Fig.~\ref{fig:fig0}.
In this paper, our \textbf{motivation} is to explore different ways to improve the dehazing performance in terms of both restoration accuracy and computational efficiency.

Despite remarkable performance of current CNN-based methods, the expressive ability (or model capacity) is still limited, which depends heavily on the feature extraction.
Our first improvement is made to enhance the feature learning ability via integrating the multi-scale scheme (in feature extraction).
During the dehazing process, the multi-scale characteristics of natural scenes are always ignored.
Since different scenes or the objects inside them have rich details and various sizes/shapes, the idea way for feature extraction should be scene/object-dependent.
However, size-fixed convolution layers are typically adopted in CNN-based dehazing methods \cite{Dong2020CVPR-MSBDN,Qin2020AAAI-FFA,Wu2021CVPR-AECR}.
Such a convolution layer with relatively fixed and single receptive field is inadequate to cover correlated areas, failing to tackle the hazy image captured under this kind of scene.
We argue that one possible solution is to utilize various scales of receptive fields in a single feature extractor.
Therefore, in this paper we propose a multi-stream feature extraction (MSFE) module which contains three parallel convolutions with different receptive fields to extract multi-scale features.
In MSFE, the large receptive field is responsible for large-scale information, e.g., dense hazy regions, while the small receptive field concentrates on fine details.
In addition, we also adopt an attention module (consisting of a channel attention and a spatial attention) to make the feature extractor adaptively focus on significant channels or regions.
The MSFE and attention modules constitute our multi-stream feature attention block (MSFAB).

The second improvement is to adapt the non-local network \cite{Wang2018CVPR-Nonlocal} to make it fit for image dehazing.
Non-local network \cite{Wang2018CVPR-Nonlocal}, which can enable the model to explore global information relationships among the whole image, has been applied to many vision tasks (e.g., super-resolution \cite{Zhang2019ICLR-RNAN,Dai2019CVPR-SAN}, semantic segmentation \cite{Huang2019CCNet}).
Although very promising results have been achieved, non-local network is seldom applied in image dehazing domain.
The main reasons behind this phenomenon are the prohibitive computational cost and vast GPU memory occupation, hindering its practice.
Therefore, how to adapt non-local network into image dehazing is a promising research direction.
We propose a cross non-local block (CNLB) to expand the search space of long-range dependencies and meanwhile simplify the matrix multiplications.
The former is achieved by exploring the similarities within and beyond the query input.
The inputs of key and value branches are no longer identical with the query, and instead more beneficial features from preceding layers are fused as the input.
The latter is realized by introducing a spatial pyramid down-sampling (SPDS) strategy.

At present, the contrastive regularization (CR) is embedded into the loss function to pull the predicted image to the clean image and push it from the hazy image (in the representation space) \cite{Wu2021CVPR-AECR}.
\hzw{Conventionally, the pre-trained VGG model \cite{Simonyan2015} is adopted for creating the representation space, which is designed for general-purpose classification.
Therefore, we re-train a VGG-haze model for distinguishing hazy scenes from haze-free.
This represents the first key difference.}
In \cite{Wu2021CVPR-AECR}, both low-level (detail information) and high-level (semantic information) feature maps are utilized to build the representation space.
However, we notice that given a certain scene, the semantic object is independent of the presence or absence of the haze.
The semantic information does not seem to help the pull or push forces.
Based on the representation space (by VGG-haze) we further present a novel detail-focused contrastive regularization (DFCR) by emphasizing the low-level details, to optimize the training direction.

Based on the above improvements (i.e., MSFAB, CNLB, DFCR), our proposed MRFNLN model outperforms existing state-of-the-art dehazing solutions \cite{Dong2020CVPR-MSBDN,Qin2020AAAI-FFA,Wu2021CVPR-AECR}, as illustrated in Fig.~\ref{fig:fig0}.
The main contributions of this paper are summarized as follows:

\begin{itemize}
	\item We design an effective local feature extraction module - multi-stream feature attention block (MSFAB), which contains three parallel convolutions with different receptive fields (i.e., $1\times1$, $3\times3$, and $5\times5$), a channel attention mechanism, and a spatial attention mechanism (with dilated convolution). This simple design can improve the expressive ability of the network by introducing multiple receptive fields and provide flexibility in dealing with various types of haze by adaptively focusing on important channels/regions.
	\item Non-local scheme is efficiently and effectively adapted to fit for image dehazing. A cross non-local block (CNLB) is proposed to expand the long-range dependencies' search space via exploring the similarities within more beneficial features. In addition, a spatial pyramid down-sampling (SPDS) strategy is introduced to mitigate the limitations of computational cost and GPU memory.
	\item We present a novel detail-focused contrastive regularization (DFCR) by emphasizing the low-level details and ignoring the high-level semantic information in a representation space specially designed for dehazing. This modified CR improves the dehazing performance without costing extra computations and parameters during the inference phase. By combining above mentioned modifications, we propose our three-level U-Net-like architecture, i.e., multi-receptive-field non-local network (MRFNLN), which achieves state-of-the-art performance among models less than \textbf{1.5 Million} parameters.
\end{itemize}


\section{Related work}
\label{sec: related work}
Recently, data-driven methods \cite{Cai2016TIP,Ren2016ECCV,Li2017ICCV-AOD,Zhang2018CVPR-DCPDN,Qin2020AAAI-FFA,Dong2020CVPR-MSBDN,Wu2021CVPR-AECR,Hong2022AAAI,Ye2022ECCV-PMDNet} have dominated this domain by achieving incredible performance.
The basic hypothesis behind these methods is that a mapping from corrupted data to ground truths or intermediate haze-related variables can be learned from substantial hazy-clean image pairs via convolutional neural networks (CNNs).
We focus on deep learning-based dehazing methods in this paper.

\subsection{Deep Image Dehazing}
With the rising of deep learning, deep dehazing models have made great progress.
Cai \etal \cite{Cai2016TIP} proposed a trainable CNN based model called DehazeNet to estimate the transmission map (i.e., $t(x)$), which is subsequently used to derive the haze-free image via ASM \cite{Nayar1999ICCV-ASM,Narasimhan2003TPAMI-ASM}.
Similarly, Ren \etal \cite{Ren2016ECCV} designed a multi-scale CNN (i.e., MSCNN) to estimate a coarse-level transmission map and later refine it to fine-level.
The global atmospheric light (i.e., $A$) is separately estimated by empirical rules for both DehazeNet and MSCNN methods.
By re-formulating the ASM, AOD-Net \cite{Li2017ICCV-AOD} unifies $t(x)$ and $A$ into one variable. 
Thus, they can be estimated simultaneously.
However, these methods may cause a cumulative error if the estimations of $t(x)$ and $A$ are inaccurate or biased, resulting in undesired artifacts and large reconstruction errors.
Besides, collecting the ground-truth of $t(x)$ is difficult or expensive in the real world.

GridDehazeNet proposed by Liu \etal \cite{Liu2019ICCV-GridDehazeNet} utilizes a grid-like CNN to directly learn hazy-to-clean image translation without referring to the ASM.
The authors claimed that directly estimating the haze-free images is better than estimating the atmospheric scattering parameters.
Following this, Dong \etal \cite{Dong2020CVPR-MSBDN} proposed a multi-scale boosted dehazing network (MSBDN) based on the U-Net architecture \cite{Ronneberger2015MICCAI-Unet}.
The decoder of MSBDN is regarded as an image restoration module and a strengthen-operate-subtract (SOS) boosting strategy is employed to progressively remove the haze.
Later, a feature fusion attention network (FFA-Net) is proposed by Qin \etal \cite{Qin2020AAAI-FFA}, which improves the performance of single image dehazing by a very large margin.
The basic module inside FFA-Net, i.e., feature attention block (FAB), treats different features and pixels unequally, and then becomes a common block in image dehazing \cite{Wu2021CVPR-AECR}.
By introducing a novel contrastive regularization (CR) to exploit both positive and negative samples, AECR ensures that the recovered image is close to the clean image and far away from the hazy image.
Hong \etal \cite{Hong2022AAAI} first took uncertainty into consideration and proposed a novel uncertainty-driven dehazing network (UDN).
Ye \etal \cite{Ye2022ECCV-PMDNet} tried to explicitly model the haze distribution via a density map and designed a separable hybrid attention (SHA) module.
Zhang \etal \cite{Zhang2022TCYB} proposed a hierarchical density-aware dehazing network to estimate a low-resolution $t(x)$ (to approximate the density information).
Recently, transformer is introduced in image dehazing.
For example, Guo \etal \cite{Guo2022CVPR-Dehamer} investigated how to combine CNN and transformer for image dehazing.
In addition, Song \etal \cite{Song2023TIP} modified the swin transformer \cite{Liu2021ICCV-swin} to make it suitable for image dehazing and pushed the state-of-the-art dehazing performance forward.
Existing deep dehazing methods mainly focus on increasing the depth and width to improve the performance.
However, the number of parameters and training difficulty will substantially increase.
In this paper, we will in turn explore efficient and effective ways.

\subsection{Non-local Network}
Non-local network is initially proposed by Wang \etal \cite{Wang2018CVPR-Nonlocal} for video classification.
Some scientists noticed that leveraging the long range dependencies brings great benefits to both low-level and high-level vision tasks \cite{Dai2019CVPR-SAN,Zhang2019ICLR-RNAN,Zhu2019ICCV-AFNB,Huang2019CCNet}.
As for image dehazing, we surprisingly find that few methods adopt non-local network.
Previous methods try to integrate the non-local conception into the loss function \cite{Zhang2020Neurocomp-NLDN} or channel attention \cite{Sun2023NN-MFIENA}.
Due to the huge computational complexity brought by the matrix multiplications, it is not easy to employ the non-local network into CNN structure, especially when the GPU memory is limited and the spatial resolution is high \cite{Qiao2020TITS-Trustworthy, Zhang2022B-WC-lockchain}.
In this paper, we adapt the non-local network to image dehazing in a more efficient way.

\begin{figure}[t]
	\centering
	\includegraphics[width=0.99\linewidth]{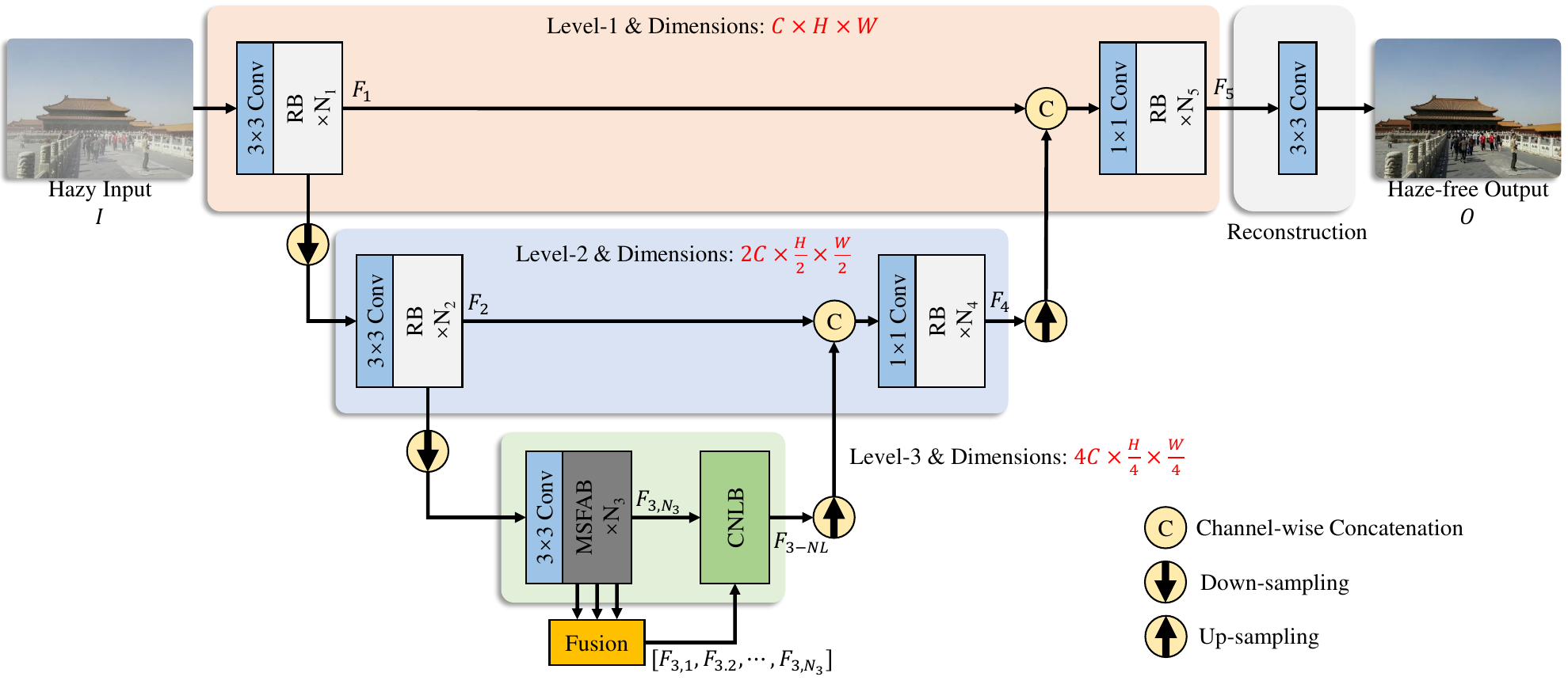}
	\caption{The overall architecture of our proposed multi-receptive-field non-local network (MRFNLN). Given a hazy image $I\in \mathbb{R}^{3\times H\times W}$ as the input, MRFNLN reconstructs its corresponding haze-free image $O\in \mathbb{R}^{3\times H\times W}$ with an end-to-end manner.}
	\label{fig:fig2}
\end{figure}

\section{Methodology}
\label{sec: methodology}

Fig.~\ref{fig:fig2} shows the overall architecture of our proposed multi-receptive-field non-local network (MRFNLN), which can be regarded as a three-level U-Net variant.
Given a hazy image $I\in \mathbb{R}^{3\times H\times W}$ as the input, MRFNLN recovers its corresponding haze-free image $O\in \mathbb{R}^{3\times H\times W}$ with an end-to-end manner.
It is a hierarchical framework with two down-sampling operations and two corresponding up-sampling operations.
The down-sampling operation halves the spatial dimensions and doubles the number of channels. 
It is realized through a normal convolution layer by setting the value of stride to 2 and setting the number of output channels to \hzw{twice} of input channels.
The up-sampling operation can be regarded as the inverse form of the down-sampling operation, which is realized through a deconvolution layer.
There are three levels in MRFNLN, and we employ different blocks in different levels to extract corresponding features.
According to \cite{Liang2021CVPR-Laplacian}, we reveal that the attribute transformations between hazy and haze-free images, such as illumination and color change, relate more to the low-frequency component (i.e., low-resolution level 3).
It is very straightforward that we deploy simple blocks in level 1 and 2, and sophisticated blocks in level 3.
Specifically, we utilize residual block (RB) \cite{He2016CVPR-ResNet}, residual block (RB), and multi-stream feature attention block (MSFAB) from level 1 to 3, respectively.
Recursive learning is adopted in these blocks
Besides, we also employ a cross non-local block (CNLB) to capture long-range dependencies in level 3.

\begin{figure}[t]
	\centering
	\includegraphics[width=0.8\linewidth]{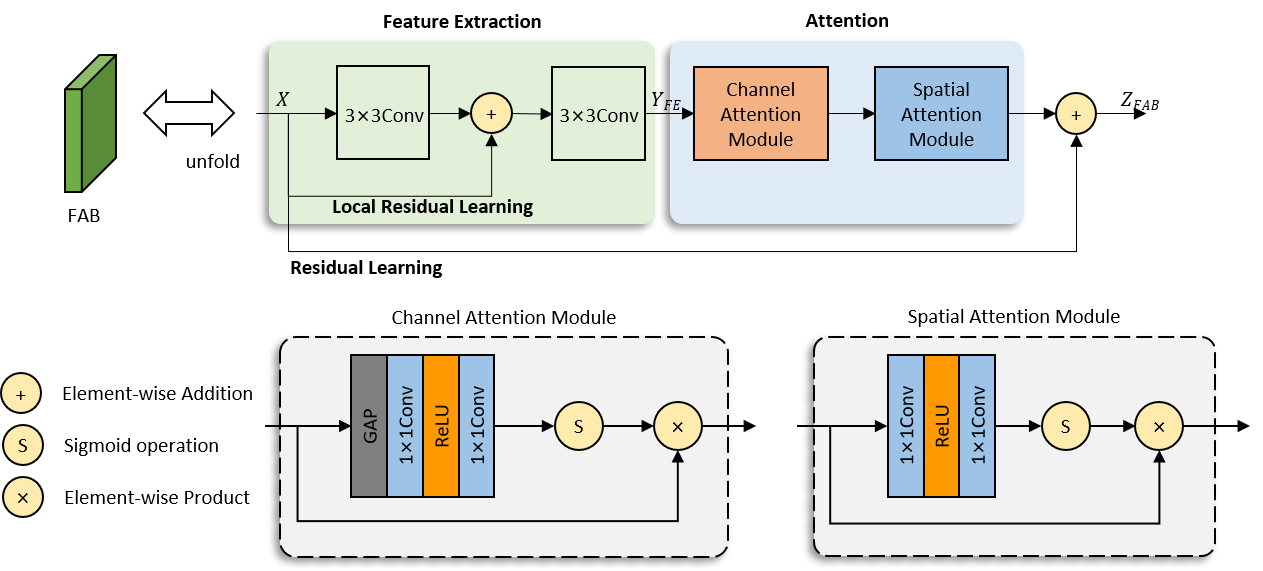}
	\caption{The detailed architecture of feature attention block (FAB) from \cite{Qin2020AAAI-FFA}. FAB contains two key parts, i.e., the feature extraction (FE) part in the light green box and the attention part in the light blue box. GAP indicates the global average pooling operation.}
	\label{fig:fig3}
\end{figure}

\begin{figure}[b]
	\centering
	\includegraphics[width=0.9\linewidth]{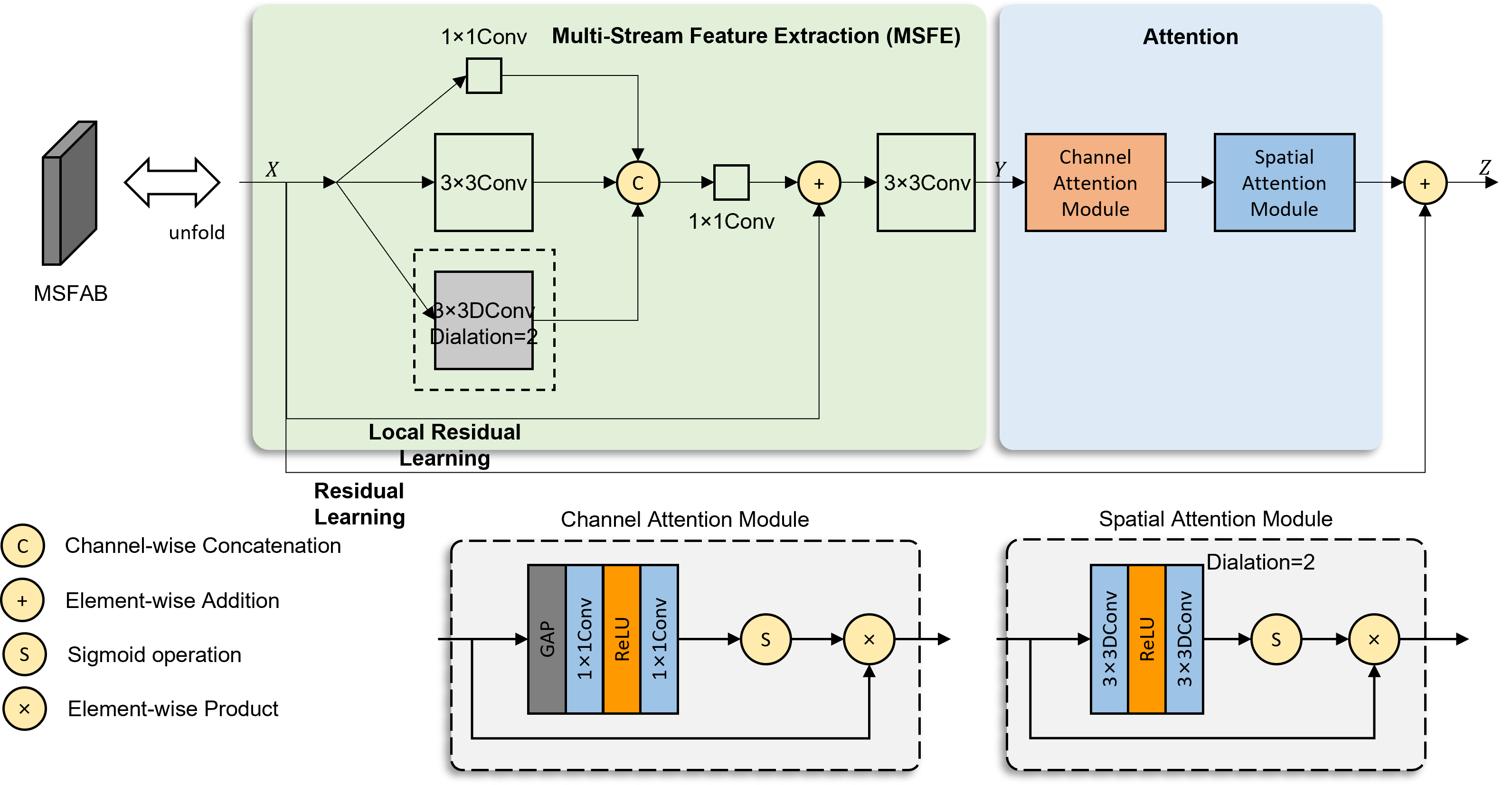}
	\caption{The detailed architecture of our proposed multi-stream feature attention block (MSFAB). MSFAB consists of two key parts, i.e., the multi-stream feature extraction (MSFE) part in the light green box and the attention part in the light blue box. DConv denotes a dilated convolution layer.}
	\label{fig:fig4}
\end{figure}

\subsection{Multi-stream Feature Attention Block}
As shown in Fig.~\ref{fig:fig3}, we first recap the feature attention block (FAB) in \cite{Qin2020AAAI-FFA}. 
It consists of feature extraction, channel attention, and spatial attention, with strong representational ability for dehazing task.
However, we argue that the disadvantage of FAB is two-fold.

On the one hand, FAB utilizes only one single convolution layer to extract feature maps, which has size-fixed receptive field.
Since receptive field is the fundamental unit for searching recovering clues in haze removing task, size-fixed receptive field is not suitable for natural hazy images with diverse patterns/details/textures.
The ideal way of reconstructing missing patterns/details/textures should be scale-dependent.
One possible solution to deal with this situation is to employ various scales of receptive fields in a single feature extractor.

As shown in Fig.~\ref{fig:fig4}, we embed a multi-stream feature extraction (MSFE) part, and replace it with the feature extract (FE) part of FAB.
Specifically, in MSFE part, we utilize a standard convolution layer using $3\times 3$ kernels, a standard convolution layer using $1\times 1$ kernels, and a dilated convolution layer using $3\times 3$ kernels with the dilation value is set to 2.
These three convolution layers are parallel deployed to extract multi-scale features with multiple receptive fields (i.e., $1\times 1$, $3\times 3$, and $5\times 5$). 
Let $X \in \mathbb{R}^{C\times H\times W}$ denote the input feature maps of MSFE part, the extracted multi-scale features $F^{1\times 1}$, $F^{3\times 3}$ and $F^{5\times 5}$ can be formulated as:
\begin{equation}
	\begin{tabular}{c}
		$F^{1\times 1} = \mathcal{C}_{1\times1}(X)$,\\
		$F^{3\times 3} = \mathcal{C}_{3\times3}(X)$,\\
		$F^{5\times 5} = \mathcal{DC}_{3\times3,dia=2}(X)$,
	\end{tabular}
\end{equation}
where $\mathcal{DC}_{k\times k,dia=d}$ denotes the dilated convolution layer using $k\times k$ kernels with the dilation value $d$.
After computing the multi-scale features, we concatenate them together channel-wisely, and then employ a $1\times 1$ convolution layer to reduce the channel number from $3C$ to $C$.
Similar to FAB, we also employ the local residual learning and a $3\times 3$ convolution layer to calculate the output of MSFE (i.e., $Y \in \mathbb{R}^{C\times H\times W}$)
The formulas are as follows:
\begin{equation}
	Y = \mathcal{C}_{3\times 3}(X + \mathcal{C}_{1\times1}([F^{1\times 1},F^{3\times 3},F^{5\times 5}])),
\end{equation}


On the other hand, the spatial attention sub-part in FAB employs two $1\times 1$ convolution layers to generate the spatial weights (see in Fig.~\ref{fig:fig3} right bottom). The output of FAB in \cite{Qin2020AAAI-FFA} (i.e., $Z_{FAB}$) can be formulated as $Z_{FAB} = X + \mathcal{F}_{SA}(\mathcal{F}_{CA}(Y_{FE})),$
where $\mathcal{F}_{SA}(\cdot)$ and $\mathcal{F}_{CA}(\cdot)$ denote the spatial attention and channel attention, respectively. 

This setting indicates that the weight in certain pixel position is only calculated based on the feature vector in this pixel position, without considering neighboring information.
Spatial importance weights calculated by this way have not been compared with neighboring pixels, which are not comprehensive enough.
Enlarging the receptive field is a simple and effective solution to encoding more neighboring features.
By following \cite{Yu2016ICLR-Dilated}, we employ two dilated $3\times 3$ convolution layers with the dilation value is set to 2 (see in Fig.~\ref{fig:fig4} right bottom).
The first dilated convolution layer reduces the channel dimension from $C$ to $\frac{C}{r}$, and the second further reduces the channel dimension from $\frac{C}{r}$ to 1 (according to \cite{Qin2020AAAI-FFA}, $r=2$).
\begin{equation}
	Z_{MSFAB} = X + \mathcal{F}_{SA\_dia}(\mathcal{F}_{CA}(Y))
\end{equation}
where $Z_{MSFAB}$ denotes the output of our proposed MSFAB, and $\mathcal{F}_{SA\_dia}(\cdot)$ denotes the spatial attention implemented by dilated convolutions.
The detailed implementation of $\mathcal{F}_{SA\_dia}(\cdot)$ is as follows:
\begin{equation}
	\begin{tabular}{c}
		$\mathcal{F}_{SA\_dia}(x) = x \times M_{SA}$ \\
		$ = x \times \sigma(\mathcal{DC}_{3\times 3,dia=2}(max(0,\mathcal{DC}_{3\times 3,dia=2}(In))))$
	\end{tabular}
\end{equation}
where $x$ denotes the input feature maps, $M_{SA\_dia}$ denotes the spatial attention weights map, $max(0,x)$ denotes the ReLU activation, and $\sigma$ indicates the \textit{Sigmoid} operation.
We will discuss the effectiveness of MSFE and dilation-based SA in Sec.~\ref{effectiveness of MSFAB}.

\subsection{Cross Non-local Scheme}
\label{subsec: CNLB}
Previous dehazing works \cite{Qin2020AAAI-FFA,Wu2021CVPR-AECR,Dong2020CVPR-MSBDN} usually employ local receptive field or deformable receptive field to exploit the information relationships in the feature space.
According to some regression tasks \cite{Berman2016,Dai2019CVPR-SAN,Liu2018NIPS-NLRN}, global receptive field is also important for mining potential information relationships (e.g., long-range correlations).  
Therefore, we try to capture long-range dependencies via introducing the non-local scheme, which is firstly proposed in \cite{Wang2018CVPR-Nonlocal} and can calculate the similarities of a certain pixel to all locations within an image.
In particular, we embed the non-local block (NLB) only in level 3 after the stacked MSFABs, since the computational cost and GPU memory occupation are mainly determined by the spatial dimensions (i.e., $H$ and $W$).

\subsubsection{Revisiting Non-local Block}

\begin{figure}[t]
	\centering
	\includegraphics[width=0.9\linewidth]{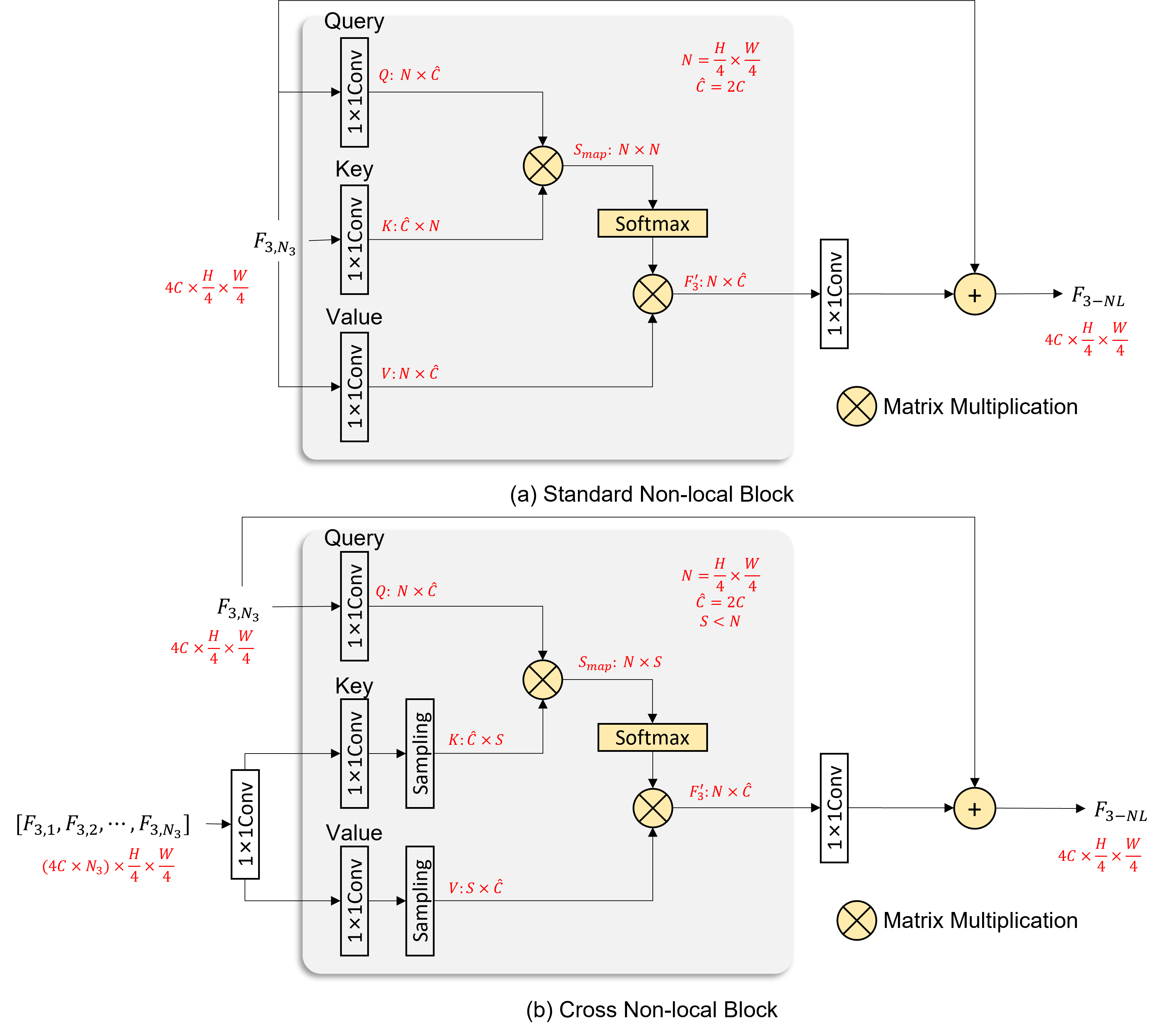}
	\caption{The detailed demonstrations of (a) the standard non-local block, and (b) proposed cross non-local block. }
	\label{fig:fig5}
\end{figure}

Fig.~\ref{fig:fig5} (a) shows the architecture of the vanilla non-local block \cite{Wang2018CVPR-Nonlocal}.
Three $1\times 1$ convolution layers $\mathcal{C}_{1\times 1}^{query}(\cdot)$, $\mathcal{C}_{1\times 1}^{key}(\cdot)$, and $\mathcal{C}_{1\times 1}^{value}(\cdot)$ are embedded to transform the output of final MSFAB in level 3 (i.e., $F_{3,N_3} \in \mathbb{R}^{4C\times \frac{H}{4} \times \frac{W}{4}}$) to corresponding embeddings $Q\in \mathbb{R}^{N \times \hat{C}}$, $K\in \mathbb{R}^{\hat{C} \times N}$, and $V\in \mathbb{R}^{N \times \hat{C}}$ (For simplification, we omit the reshape operations following these three branches.).
\begin{equation}
	Q = \mathcal{C}_{1\times 1}^{query}(F_{3,N_3}), K = \mathcal{C}_{1\times 1}^{key}(F_{3,N_3}), V = \mathcal{C}_{1\times 1}^{value}(F_{3,N_3}),
	\label{Eqn14}
\end{equation}
where $\hat{C}$ denotes the channel number of the obtained embedding, and $N$ presents the total number of spatial locations. In our implementation, we set $\hat{C} = 2C$ and $N=\frac{H}{4} \times \frac{W}{4}$.
Then, we compute the similarity matrix $S_{map} \in \mathbb{R}^{N\times N}$ via a matrix multiplication operation.
\begin{equation}
	S_{map} = Q \times K,
	\label{Eqn15}
\end{equation}
Afterward, the matrix is normalized by a \emph{Softmax} operation. 
Another $1\times 1$ convolution layer is employed to act as a weighting parameter to adjust the importance of the non-local operation \emph{w.r.t.} the original input \cite{Zhu2019ICCV-AFNB}, and moreover, expand the channel number back to $4C$ from $\hat{C}$.
Formally, the NLB is defined as:
\begin{equation}
	F_{3-NL} = \mathcal{C}_{1\times 1}(\Gamma(Softmax(S_{map}) \times V)) + F_{3,N_3},
	\label{Eqn16}
\end{equation}
where $\Gamma(\cdot)$ indicates the reshape operation to transform the dimensions from $N \times \hat{C}$ to $\hat{C} \times \frac{H}{4} \times \frac{W}{4}$. The output feature $F_{3-NL} \in \mathbb{R}^{4C\times \frac{H}{4} \times \frac{W}{4}}$ is refined with all locations in $F_{3,N_3}$, enabling it with the global receptive field.

\subsubsection{Cross Non-local Block}
Although, standard NLB is proved to work well in many tasks \cite{Zhang2019ICLR-RNAN,Liu2018NIPS-NLRN}, there are still two limitations.
(1) It can only capture the long-range dependencies within the input features (i.e., $F_{3,N_3}$).
(2) It is also criticized for prohibitive computational cost and GPU memory usage.

\textbf{In order to tackle the first limitation}, we try to explore the long-range dependencies beyond the query input self.
The standard NLB has only one input source, which means the query, key, and value branches are based on the same features.
As shown in Fig.~\ref{fig:fig5} (b), we provide an alternative that calculates the correlations between every pixel of $F_{3,N_3}$ and all preceding features in level 3, called cross non-local block (CNLB).
Specifically, we fuse all preceding features in level 3 by concatenating them along channel dimension to produce $[F_{3,1},F_{3,2},\cdots,F_{3,N_3}] \in \mathbb{R}^{(4C\times N_3)\times \frac{H}{4} \times \frac{W}{4}}$.
A $1\times 1$ convolution is further employed to reduce the channel dimension and generate $F_f \in \mathbb{R}^{4C \times \frac{H}{4} \times \frac{W}{4}}$.
Accordingly, we re-write Eqn.~\ref{Eqn14} as:
\begin{equation}
	Q = \mathcal{C}_{1\times 1}^{query}(F_{3,N_3}), K = \mathcal{C}_{1\times 1}^{key}(F_{f}), V = \mathcal{C}_{1\times 1}^{value}(F_{f}),
	\label{Eqn17}
\end{equation}

The proposed CNLB attempts to compute the correlations between every pixel of $F_{3,N_3}$ and $F_f$, which implies expanding the search region of NLB from one single feature map to multiple feature maps (fused version).
Therefore, CNLB can provide more sufficient dependencies than standard NLB.

\begin{figure}[h]
	\centering
	\includegraphics[width=0.75\linewidth]{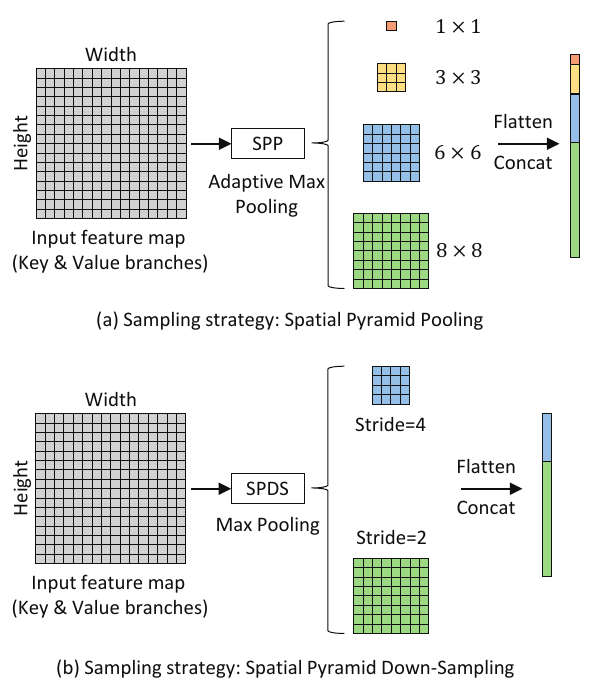}
	\caption{The detailed demonstrations of (a) spatial pyramid pooling, and (b) spatial pyramid down-sampling.}
	\label{fig:fig6}
\end{figure}

\textbf{As for the second limitation}, since two matrix multiplications in Eqn.~\ref{Eqn14} and Eqn.~\ref{Eqn15} are the main cause of the inefficiency, we sample a few representative points from key branch and value branch to directly simplify the calculation process.
Our initial idea is originated from \cite{Zhu2019ICCV-AFNB}, which employs spatial pyramid pooling (SPP) to largely reduce the computational overhead of matrix multiplications yet provide substantial feature statistics with applications to semantic segmentation.
It is clearly depicted in Fig.~\ref{fig:fig6} (a), where four adaptive max pooling layers\footnote{Different from normal pooling layer, adaptive pooling layer can automatically choose the values of stride and kernel size by calculating from input size and user-defined output size, and use them to produce output of the desired size.} are utilized after $\mathcal{C}_{1\times 1}^{key}(\cdot)$ or $\mathcal{C}_{1\times 1}^{value}(\cdot)$ and then the four pooling results are flattened and concatenated to generate the embeddings.
However, differs from segmentation, image dehazing task needs to densely predict the haze-free output.
Feature statistics (semantic level information) generated by SPP offer limited assistance to the recovery process.

Considering this, we propose the spatial pyramid down-sampling (SPDS) to reserve the contextual information meanwhile reduce the computational cost and GPU memory occupation.
As shown in Fig.~\ref{fig:fig6} (b), we adopt two max pooling layers (with different strides and kernel sizes) in key and value branches to down-sample the input feature map.
Similarly, the down-sampled feature maps are flattened and concatenated to generate the embeddings (i.e., $K \in \mathbb{R}^{\hat{C} \times S}$ and $V \in \mathbb{R}^{S\times \hat{C}}$).
In our model, we set stride = kernel size $=\{2, 4\}$, and thus the $S=\frac{HW}{8^2}+\frac{HW}{16^2}$.
Accordingly, we further re-write Eqn.~\ref{Eqn17} as:
\begin{equation}
	Q = \mathcal{C}_{1\times 1}^{query}(F_{3,N_3}), K = \mathcal{S}(\mathcal{C}_{1\times 1}^{key}(F_{f})), V = \mathcal{S}(\mathcal{C}_{1\times 1}^{value}(F_{f})),
\end{equation}
where $\mathcal{S}(\cdot)$ denotes the sampling operation.
In this situation, the spatial size of similarity matrix calculated by Eqn.~\ref{Eqn15} decreases from $N\times N$ to $N\times S$.
As a consequence, the complexity of matrix multiplication in our proposed CNLB is only $\frac{S}{N}=0.3125$ times of the complexity of matrix multiplication in NLB.
We will discuss the effectiveness of our proposed CNLB in Sec.~\ref{effectiveness of CNLB}.



\subsection{Novel Detail-focused Contrastive Regularization}
Traditionally, deep learning-based dehazing methods \cite{Qin2020AAAI-FFA,Dong2020CVPR-MSBDN,Li2017ICCV-AOD} employ positive-orient loss functions (e.g., mean absolute error, mean square error) to drive the network learning.
Among these methods, only positive samples (i.e., clean images or ground truth) are used as upper bound to guide the dehazing process \cite{Wu2021CVPR-AECR}.
Recently, some approaches try to adopt contrastive regularization in the reconstruction loss to further improve the dehazing performance. 
AECR-Net \cite{Wu2021CVPR-AECR} is a very representative work, which exploits the information of hazy images and haze-free clean images as the negative and positive samples, respectively.
By combining L1 reconstruction loss with contrastive regularization, AECR-Net can pull the recovered image (i.e., anchor) to the clean image (i.e., positive), meanwhile push the recovered image from the hazy image (i.e., negative).
We follow the AECR-Net profile and propose our detail-focused contrastive regularization (DFCR).

\hzw{Conventionally, the pre-trained VGG model \cite{Simonyan2015} is adopted for creating the representation space, which is designed for general-purpose classification.
We re-train a VGG-haze model for the binary classification of distinguishing between hazy and haze-free images.}
We take the input of MRFNLN (i.e., $I$) and the output of MRFNLN (i.e., $O$) as the `negative' pair. 
Similarly, the `positive' pair consists of the clean ground truth (i.e., $J$) and the $O$.
The contrastive regularization item can be formulated as:
\begin{equation}
	\mathcal{L}_{CR} = \sum_{i=1}^{n} w_i\cdot \frac{L_1(VGGhaze_i(J),VGGhaze_i(O))}{L_1(VGGhaze_i(I),VGGhaze_i(O))},
\end{equation}
where $VGGhaze_i(\cdot)$ extracts the $i$-th intermediate feature maps from the pre-trained VGG-haze model, $L_1(a,b)$ calculates the L1 distance between $a$ and $b$, and $w_i$ denotes the weight coefficient of $i$-th item.

Further, we notice that given a certain image, the semantic object is independent of the presence or absence of the haze.
For example, imagine a hazy scene with a sedan inside, the semantic information (i.e., the object category - sedan) will not change no matter the existence of haze or not.
In image dehazing, the low-level details encoded in shallow features are more relevant to the haze than the high-level semantic information encoded in deep features.

Therefore, we present a novel detail-focused contrastive regularization (DFCR) by emphasizing the low-level details and ignoring the high-level semantic information.
Specifically, we select only the 1st, 3rd, 5th layers to construct the feature space, and the corresponding weight coefficients $w_i$ are set to 1 for these layers.
This weight setting makes sense because we argue that the shallow layer is more important than the deep one.
We will discuss the effectiveness of DFCR in Sec.~\ref{effectiveness of DFCR}.

\section{Experiments}
\label{sec: experiment}

\subsection{Experimental Configuration}
\noindent \textbf{Datasets.}
Since collecting a large number of real-world hazy-clean image pairs is impractical, we train our MRFNLN on synthetic datasets.
REalistic Single Image DEhazing (RESIDE) \cite{Li2018TIP-RESIDE} is a widely-used dataset, which contains five subsets: Indoor Training Set (ITS), Outdoor Training Set (OTS), Synthetic Objective Testing Set (SOTS), Real-world Task-driven Testing Set (RTTS), and Hybrid Subjective Testing Set (HSTS).
We select ITS and OTS in the training phase and select SOTS in the testing phase.
Note that, the SOTS is divided into two subsets (i.e., SOTS-indoor and SOTS-outdoor) for evaluating the models separately trained on ITS and OTS.
ITS contains 1399 indoor clean images and for every clean image, 10 simulated hazy images are generated based on the physical scattering model with different parameters.
As for OTS, we pick around 294,980 images for the training process\footnote{Following \cite{Liu2019ICCV-GridDehazeNet}, data cleaning is applied since the intersection of training and testing images. Besides, some small-sized images are also removed.}.
SOTS-indoor and SOTS-outdoor contain 500 indoor and 500 outdoor testing images, respectively.
In addition, Haze4K dataset \cite{Liu2021ACMMM-Haze4K}, which contains 3000 synthetic training images and 1000 synthetic testing images, is also employed to further verify the effectiveness of our proposed MRFNLN.

\begin{table}[h]
	\footnotesize
	\centering
	\caption{Ablation study of our proposed MSFAB with different architectures. The PSNR values are tested on SOTS-indoor dataset.}
	\label{Tab1}
	\begin{tabular}{l|cccc|c}
		\toprule
		Model                 &FE &MSFE &CA+SA & CA+SA\_{dia}& PSNR (dB)   \\
		\midrule
		RB &&&&& 34.33\\
		\midrule
		FAB (Baseline)   &$\checkmark$& &$\checkmark$ && 36.23  \\
		FE$\rightarrow$MSFE &&$\checkmark$&$\checkmark$&& 37.19\\
		FE$\rightarrow$parallel FE &$\checkmark$&&$\checkmark$&& 36.30\\
		MSFAB                 &&$\checkmark$&&$\checkmark$& 37.89    \\
		\bottomrule
	\end{tabular}
\end{table}

\noindent \textbf{Evaluation Metrics.}
Peak signal-to-noise-ratio (PSNR) and structural similarity index (SSIM) \cite{Wang2004TIP-SSIM}, which are commonly used to measure the image quality among the computer vision community, are utilized for dehazing performance evaluation.
For a fair comparison, we calculate the metrics based on the RGB color images without cropping pixels.

\noindent \textbf{Implementation Details.}
We implement the proposed MRFNLN model on PyTorch deep learning platform with a single NVIDIA RTX4090 GPU.
We deploy RB, RB, and MSFAB in level 1, level 2, and level 3, respectively, and set channel number $C=32$.
The MRFNLN is optimized using Adam \cite{Kingma2015ICLR-Adam} optimizer with default hyper-parameters.
Moreover, the initial learning rate and the batch size are set to $2e^{-4}$ and $16$, respectively.
During the training, cosine annealing strategy \cite{He2019CVPR-Bag} is adopted to adjust the learning rate from the initial value to $1e^{-6}$.
The total number of training iterations on ITS, OTS and Haze4K is set to 1,500K around.
During training, we randomly crop patches from the original images, and then two data augmentation techniques are adopted including: random rotation and vertical/horizontal flip.
In our work, two MRFNLN variants are provided (MRFNLN-B and MRFNLN-L for basic and large, respectively).
For MRFNLN-B, the number of blocks deployed on different stages $[N_1, N_2, N_3, N_4, N_5]$ is set to $[1, 2, 4, 2, 1]$.
For MRFNLN-L, $[N_1, N_2, N_3, N_4, N_5]$ is set to $[2, 4, 8, 4, 2]$.

\subsection{Ablation Study}
To demonstrate the effectiveness of our multi-receptive-field non-local network (MRFNLN), we perform ablation study to verify the contribution of each component, including (1) multi-stream feature attention block (MSFAB), (2) cross non-local block (CNLB), and (3) detail-focused contrastive regularization (DFCR).

\subsubsection{The effectiveness of MSFAB}
\label{effectiveness of MSFAB}
Feature attention block (FAB), initially proposed in \cite{Qin2020AAAI-FFA}, treats different features and pixels unequally, which can provide additional flexibility in dealing with different types of information. Afterward, some deep learning-based dehazing approaches directly adopt FAB as a basic module \cite{Wu2021CVPR-AECR}, and achieve promising results. In our experiments, we also choose FAB from \cite{Qin2020AAAI-FFA} as our baseline block in level 3.

Subsequently, we modify the baseline by introducing some new features as: (1) \textbf{FE$\rightarrow$MSFE}: replace the feature extraction part in the baseline with the multi-stream feature extraction (MSFE) and keep the attention part unchanged, (2) \textbf{FE$\rightarrow$parallel FE}: deploy three parallel convolutions with the same receptive field (i.e., $3\times 3$) in the feature extraction part in the baseline, (3) \textbf{MSFAB}: introduce dilated convolutions into the spatial attention sub-part of (1) to expand the receptive field when generating the spatial weights.
These blocks mentioned above are tested in level 3, and the results are shown in Table~\ref{Tab1}.

\begin{table*}[t]
	\footnotesize
	\centering
	\caption{Ablation study of our proposed CNLB with different designs. We systematically analyze the effectiveness of the components inside CNLB. The evaluation metrics are measured on SOTS-indoor dataset.}
	\label{tab:table3}
	\begin{tabular}{c|c|cc|ccccc}
		\toprule
		\multicolumn{2}{c|}{Model}         &model-Base&model-A& Base+NL & A+NL     & A+CNL      & A+CNL{\tiny SPP} & A+CNL{\tiny SPDS}    \\
		\midrule
		\midrule
		\multirow{3}{*}{Setting} & Level 1 &RB&RB  & RB         & RB          & RB &RB &RB\\
		& Level 2 &RB&RB  & RB  & RB  & RB & RB&RB\\
		& Level 3 &FAB&MSFAB & FAB+NL& MSFAB+NL & MSFAB+CNL  &MSFAB+CNL{\tiny SPP} &MSFAB+CNL{\tiny SPDS}\\
		\midrule
		\multicolumn{2}{c|}{PSNR (dB)}     &36.23&37.89 & 36.35& 38.18      &  38.37      & 37.94  & 38.38  \\
		\multicolumn{2}{c|}{GPU memory}    &~5GB&~5GB & ~11GB & ~11GB      &~11GB       &  ~6GB      &~6GB\\
		\multicolumn{2}{c|}{\#Param.}      &861,091 &1,097,300 &894,179 &1,130,388   & 1,196,052      &1,196,052   & 1,196,052      \\
		\bottomrule
	\end{tabular}
\end{table*}

\begin{table}[t]
	\footnotesize
	\centering
	\caption{Ablation study of our proposed DFCR. $^\ast$ indicates that we re-train the AECR model according to the details in \cite{Wu2021CVPR-AECR} and the public source codes.}
	\label{tab:CR ablation}
	\begin{tabular}{l|cc}
		\toprule
		Model                                  &  A+CNL{\tiny SPDS}    & AECR-Net$^\ast$ \\
		\midrule \midrule
		w/o CR (Baseline)                      & 38.38 dB &   35.86   \\
		\midrule
		w/ original CR from AECR-Net & 39.59 dB & 37.01      \\
		w/ SIFCR (VGG-19) &38.65 dB & 36.01\\ 
		\midrule
		w/ DFCR (VGG-19)               & 39.98 dB & 37.50      \\
		w/ DFCR (VGG-haze)              & 40.22 dB & 37.70      \\
		\bottomrule
	\end{tabular}
\end{table}

For fair comparison, all of the experiments are conducted by using MRFNLN-B structure without non-local scheme and contrastive regularization (CR).
We only change the blocks used in level 3 to eliminate the impact from other factors.
For convenience, we train the models for only 750K iterations.
Although these values are lower than the fully trained models reported in Table~\ref{tab:benchmark}, these values and trends are consistent and meaningful.

The performance of aforementioned models is summarized in Table~\ref{Tab1}.
Employing MSFE brings 0.96 dB improvement on SOTS-indoor.
One may doubt if the improvement is obtained by the increased parameters, we also dig into this question.
We notice that parallel FE can bring limited improvement (only 0.07 dB) with more parameters than FAB and MSFE.
These results indicate that extracting features with multi-scale receptive fields can definitely boost the recovery accuracy.

By further modifying the spatial attention sub-part, our MSFAB outperforms the alternatives with 37.89 dB.
Enlarging the receptive field can encode more neighboring features to help generate the spatial importance weights. 

We denote the baseline and best-performance model as \textbf{model-Base} and \textbf{model-A} respectively.
Note that, we can deploy the proposed MSFAB in level 2 or level 1 to further promote the dehazing performance.
However, it will introduce more parameters and extra computational cost.
%
%


By considering the trade-off between performance and efficiency, we choose \textbf{model-A} in the following experiments for our implementation.

\subsubsection{The effectiveness of CNLB}
\label{effectiveness of CNLB}
Non-local block (NLB) \cite{Wang2018CVPR-Nonlocal} can capture long-range dependencies which are crucial for some image restoration tasks \cite{Liu2018NIPS-NLRN, Zhang2019ICLR-RNAN}.
In our work, we directly apply original NLB on our \textbf{model-Base} and \textbf{model-A} (in level 3 after the final FAB/MSFAB\footnote{We choose to not deploy NLB in level 1\&2, since the limited GPU memory.}).
We observe robust performance improvements for both models in Table~\ref{tab:table3} (adding original NLB on \textbf{model-Base/model-A} brings 0.12 dB/0.29 dB improvements.). 
Our experiments verify the effectiveness of NLB.
Very interestingly, adding NLB on \textbf{model-A} (i.e., \textbf{A+NL}) obtains more performance gains than adding NLB on \textbf{model-Base} (i.e., \textbf{Base+NL}), which also indicates that MSFAB extracts richer features than FAB \cite{Dong2020CVPR-MSBDN}. 
By searching the latent correlations within richer features, more effective long-range dependencies can be mined.

Then, we further propose the cross non-local block (CNLB) to address the limitations of NLB.
We first change the input features of the key and value branches from $F_{3,N_3}$ to $F_f$, and we denote this model as \textbf{A+CNL}.
As shown in Table~\ref{tab:table3}, \textbf{A+CNL} model outperforms \textbf{model-A} by 0.48 dB, and meanwhile achieves better performance than \textbf{A+NL}.
We argue the main reason for the high effectiveness of CNLB is that it can expand the search region from one single feature map to multiple feature maps for mining substantial latent correlations.
More long-range dependencies may bring more sufficient haze removal clues, generating clearer haze-free outputs.

\begin{table*}[t]
	\footnotesize
	\centering
	\caption{Quantitative comparisons between our proposed MRFNLN models and some state-of-the-art dehazing methods on SOTS-indoor, SOTS-ourdoor, and Haze4K datasets. We report PSNR, SSIM, number of parameters (\# Param.), number of floating-point operations (\# FLOPs) to perform comprehensive comparisons. The sign ``-'' denotes the digit is unavailable. \textbf{Bold} and \underline{underlined} indicate the best and the second best performance, respectively.}
	\label{tab:benchmark}
	\begin{tabular}{l|cc|cc|cc|cc}
		\toprule
		\multirow{2}{*}{Method} & \multicolumn{2}{c|}{SOTS-indoor} & \multicolumn{2}{c|}{SOTS-outdoor} & \multicolumn{2}{c|}{Haze4K \cite{Liu2021ACMMM-Haze4K}} & \multicolumn{2}{c}{Overhead}\\
		& PSNR & SSIM & PSNR & SSIM & PSNR & SSIM & \# Param. (M) & \# FLOPs (G)  \\
		\midrule
		\midrule
		(TPAMI'10) DCP \cite{He2011TPAMI-DCP} & 16.61 & 0.8546 & 19.14 & 0.8605 & 14.01 & 0.76 & - & -  \\
		(TIP'16) DehazeNet \cite{Cai2016TIP} & 19.82 & 0.8209 & 27.75 & 0.9269 & 19.12 & 0.84 & 0.008 & 0.5409 \\
		(ICCV'17) AOD-Net \cite{Li2017ICCV-AOD} & 20.51 & 0.8162 & 24.14 & 0.9198 & 17.15 & 0.83 & 0.0018 & 0.1146  \\
		(CVPR'18) GFN \cite{Ren2018CVPR-GFN} & 22.30 & 0.8800 & 21.55 & 0.8444 & - & - & 0.4990 & 14.94  \\
		(ICCV'19) GridDehazeNet \cite{Liu2019ICCV-GridDehazeNet} & 32.16 & 0.9836 & 30.86 & 0.9819 & 23.29 & 0.93 & 0.9557 & 18.71 \\
		\midrule
		(AAAI'20) FFA-Net \cite{Qin2020AAAI-FFA} & 36.39 & 0.9886 & 33.57 & 0.9840 & 26.97 & 0.95 & 4.456 & 287.5 \\
		(CVPR'20) MSBDN \cite{Dong2020CVPR-MSBDN} & 32.77 & 0.9812 & 34.81 & 0.9857 & 22.99 & 0.85 & 31.35 & 41.54  \\
		(ACMMM'21) DMT-Net \cite{Liu2021ACMMM-Haze4K} & - & - & - & - & 28.53 & 0.96 & 51.79 & 75.56  \\
		(CVPR'21) AECR-Net \cite{Wu2021CVPR-AECR} & 37.17 & 0.9901 & - & - & - & - & 2.611 & 52.20 \\
		(TIP'22) SGID-PFF \cite{Bai2022TIP-SGID} & 38.52 & 0.9913 & 30.20 & 0.9754 & - & - & 13.87 & 152.8  \\
		(AAAI'22) UDN \cite{Hong2022AAAI} & 38.62 & 0.9909 & 34.92 & 0.9871 & - & - & 4.250 & -  \\
		(ECCV'22) PMDNet \cite{Ye2022ECCV-PMDNet} & 38.41 & 0.9900 & 34.74 & 0.9850 & 33.49 & 0.98 & 18.90 & -  \\
		(CVPR'22) Dehamer \cite{Guo2022CVPR-Dehamer} & 36.63 & 0.9881 & 35.18 & 0.9860 & - & - & 132.4 & 48.93  \\
		(CVPR'22) MAXIM \cite{Tu2022CVPR-MAXIM} & 38.11 & 0.9910 & 34.19 & 0.9850 & - & - & 13.35 & 206.7 \\
		(TIP'23) Dehazeformer-M \cite{Song2023TIP} &38.46 & 0.9940 & 34.29 & 0.9830 & - & - & 4.634 & 48.64 \\
		(CVPR'23) C2PNet \cite{Zheng2023CVPR-C2PNet} &\underline{42.56} & 0.9954 & \underline{36.68} & \underline{0.9900} & - & - & -& -\\
		(TIP'24) DEA-Net-CR \cite{Chen2024TIP-DEA-Net} & 41.31 & 0.9945 & 36.59 & 0.9897 & 34.25 & 0.99 & 3.653 & 32.23 \\
		(TMM'24) PNE \cite{Cheng2024TMM} & 42.20 & \underline{0.9955} & - & - & 31.07 & 0.98 & 4.750 & - \\
		\midrule
		(Ours) MRFNLN-B &41.04&0.9948& 36.33 &0.9899&\underline{34.26}&\underline{0.99}&1.196&19.03\\
		(Ours) MRFNLN-L &\textbf{42.61} & \textbf{0.9960}&\textbf{37.68} &\textbf{0.9920} & \textbf{34.78}& \textbf{0.99}&1.262&33.74\\
		\bottomrule
	\end{tabular}
\end{table*}

However, both NLB and CNLB are very time and memory consuming compared with normal operations in deep learning, e.g., activation and convolution.
When comparing with \textbf{model-A}, \textbf{A+CNL} occupies round two times of GPU memory (5GB \textit{vs.} 11GB).
It is worth mentioning that the digits are measured with only one non-local block in level 3.
The NLB/CNLB is very unfriendly to GPUs with limited memory.
As shown in Table~\ref{tab:table3}, employing proposed spatial pyramid down-sampling (SPDS) into \textbf{A+CNL} (denoted as \textbf{A+CNL{\tiny SPDS}}) can effectively reduce the GPU memory usage (11GB $\rightarrow$ 6GB) without sacrificing the performance (the PSNR value even increases by 0.01 dB).
In addition, we also compare the spatial pyramid pooling (SPP) strategy (denoted as \textbf{A+CNL{\tiny SPP}}) in Table~\ref{tab:table3}.
The results indicate that semantic level information may damage/harm the haze removal process.

Based on the above analysis, we select \textbf{A+CNL{\tiny SPDS}} in the following experiments.

\begin{figure*}[t]
	\centering
	\includegraphics[width=0.99\linewidth]{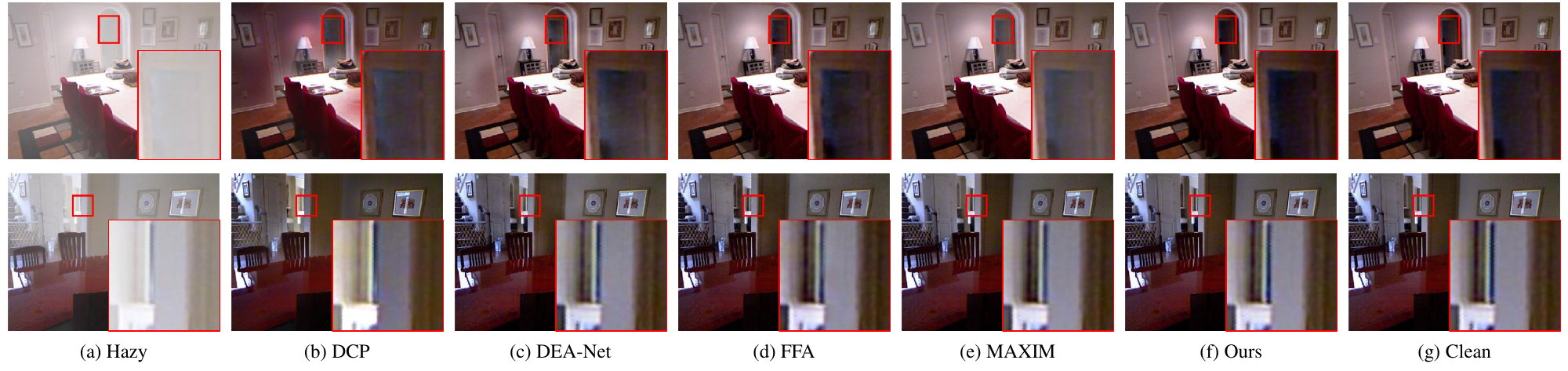}
	\caption{Visual comparisons of various methods on synthetic SOTS-indoor \cite{Li2018TIP-RESIDE} dataset. Please zoom in on screen for a better view.}
	\label{fig:Comparison_ITS}
\end{figure*}

\begin{figure*}[!h]
	\centering
	\includegraphics[width=0.99\linewidth]{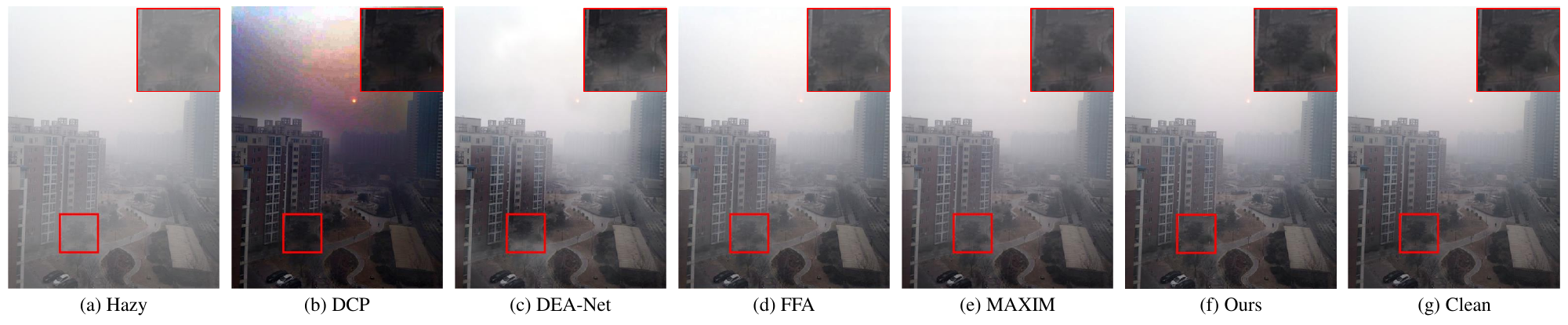}
	\caption{Visual comparisons of various methods on synthetic SOTS-outdoor \cite{Li2018TIP-RESIDE} dataset. Please zoom in on screen for a better view.}
	\label{fig:Comparison_OTS}
\end{figure*}

\subsubsection{The effectiveness of DFCR}
\label{effectiveness of DFCR}
Then, we investigate the effectiveness of the novel detail-focused contrastive regularization (DFCR).
We compare DFCR with different settings on \textbf{A+CNL{\tiny SPDS}} and AECR-Net \cite{Wu2021CVPR-AECR}.
As shown in Table~\ref{tab:CR ablation}, CR can robustly improve the performance by over 1 dB, which is not marginal in dehazing domain.
We can also observe that by emphasizing the low-level details in the representation space, our DFCR achieves better performance on both network structures, promoting the PSNR metric by over 1.5 dB against the baseline.
In addition, since DFCR only needs to extract the low-level features to create the representation space, it occupies less GPU memory during the training phase than original CR \cite{Wu2021CVPR-AECR}.

The original CR extracts low-level details and high-level semantic information simultaneously.
However, given a certain image, whether it contains haze or not, the semantic object is static and will not change.
In image dehazing, the low-level details are more relevant to the haze than the high-level semantic information.
That explains the superiority of DFCR against original CR.
Note that, our DFCR costs no extra computations and parameters during the inference phase.

We also employ a semantic information-focused contrastive regularization (SIFCR), which emphasizing the high-level semantic information, to implement the contrastive learning.
In SIFCR, only 9th and 13th layers of VGG-19 are selected to generate the feature space.
As depicted in Table~\ref{tab:CR ablation}, SIFCR brings incremental performance improvement when compared with DFCR, which further validates our hypothesis.
\hzw{Finally, using re-trained VGG-haze model to create the representation space brings further improvements.}

\subsection{Comparisons with SOTA methods}

In this section, we compare our MRFNLN with 5 early dehazing approaches including DCP \cite{He2011TPAMI-DCP}, DehazeNet \cite{Cai2016TIP}, AOD-Net \cite{Li2017ICCV-AOD}, GFN \cite{Ren2018CVPR-GFN}, GridDehazeNet \cite{Liu2019ICCV-GridDehazeNet} and 13 recent state-of-the-art (SOTA) deep dehazing methods including FFA-Net \cite{Qin2020AAAI-FFA}, MSBDN \cite{Dong2020CVPR-MSBDN}, DMT-Net \cite{Liu2021ACMMM-Haze4K}, AECR-Net \cite{Wu2021CVPR-AECR}, SGID-PFF \cite{Bai2022TIP-SGID}, UDN \cite{Hong2022AAAI}, PMDNet \cite{Ye2022ECCV-PMDNet}, Dehamer \cite{Guo2022CVPR-Dehamer}, MAXIM \cite{Tu2022CVPR-MAXIM}, C2PNet \cite{Zheng2023CVPR-C2PNet}, Dehazeformer \cite{Song2023TIP}, DEA-Net \cite{Chen2024TIP-DEA-Net}, PNE \cite{Cheng2024TMM} on SOTS-Indoor, SOTS-Ourdoor, and Haze4K datasets.
Their evaluation metrics are obtained by using their official codes or from published papers if they are available, otherwise we re-trained the models using the same training datasets.

\subsubsection{Quantitative Comparison}
Table~\ref{tab:benchmark} reports the average PSNR and SSIM values of the competitors on SOTS and Haze4K datasets.
We observe that even the basic MRFNLN-B model ranks relatively high in terms of PSNR and SSIM.
The large model MRFNLN-L outperforms the competitors on all three datasets.

In addition, we utilize number of parameters (\# Param.), number of floating-point operations (\# FLOPs) to indicate competitors' computational efficiencies.
Except the early dehazing methods, our MRFNLN models are compact in terms of parameter size.
Similar comparative results are observed in terms of FLOPs.
The \# FLOPs are measured on a color image with a resolution of $256\times 256$.

It is worth mentioning that our MRFNLN-B/-L model achieves the state-of-the-art performance on SOTS (including -indoor and -outdoor) and Haze4K datasets with less than 1.5 Million parameters.


\subsubsection{Qualitative Comparison}

Fig.~\ref{fig:Comparison_ITS} visualizes the recovered images of our MRFNLN and previous SOTA methods on synthetic SOTS-indoor dataset.
It can be observed that DCP method suffers from severe color distortion and artifacts.
The results of the competitors contain obvious haze residues.
Instead, our proposed MRFNLN model generates more natural restoration results, preserving sharper and clearer contours or edges.  
Similarly, Fig.~\ref{fig:Comparison_OTS} visualizes the recovered images from synthetic SOTS-outdoor dataset by different methods.
The DCP fails to suppress artifacts in the sky region.
We notice that in outdoor scenes, the result of our MRFNLN model is closest to the ground truth than the other alternatives.

\section{Conclusion}
\label{sec: conclusion}

In this paper, we propose a MRFNLN model for image dehazing. The design considers both reconstruction accuracy and computational efficiency. First, a MSFAB is introduced to extract multi-scale features, leveraging multi-receptive-field profiles for recovery clues. Second, we adapt the non-local block for dehazing by expanding the search space for long-range dependencies and reducing computations, resulting in a CNLB with a SPDS strategy to simplify matrix multiplications. Finally, a DFCR is integrated into the loss function to improve optimization in the representation space. Experimental results validate the efficiency and effectiveness of MRFNLN.


\newpage
\bibliographystyle{IEEEtran}
\bibliography{reference}

\vfill

\end{document}